%% file: acra.tex
\documentclass{article}
\usepackage{acra}
\usepackage{times}
\usepackage{epsfig}
\usepackage{graphicx}
\usepackage{amsmath}
\usepackage{amssymb}
\usepackage{adjustbox}
\usepackage{pifont}
\newcommand{\cmark}{\ding{51}}
\newcommand{\xmark}{\ding{55}}
\usepackage{tabularx}

\title{Point Cloud Segmentation Using Sparse Temporal Local Attention}
\author{Joshua Knights$^{1,2}$, Peyman Moghadam$^{1,2}$, Clinton Fookes$^{2}$, Sridha Sridharan$^{2}$ \\ $^{1}$ Robotics and Autonomous Systems, CSIRO, Data61, $^{2}$ Queensland University of Technology   \\ 
joshua.knights@csiro.au  peyman.moghadam@csiro.au \\ c.fookes@qut.edu.au s.sridharan@qut.edu.au}

\begin{document}

\maketitle

\input{Chapters/00_abstract.tex}
\input{Chapters/01_Introduction}

\input{Chapters/02_Related_Works}

\input{Chapters/03_Methodology}

\input{Chapters/04_Experiments}

\input{Chapters/05_Conclusion}

{\small
\bibliographystyle{apalike}
\bibliography{egbib}
}

\end{document}

%% file: Chapters/00_abstract.tex
\begin{abstract}
Point clouds are a key modality used for perception in autonomous vehicles, providing the means for a robust geometric understanding of the surrounding environment.  However despite the sensor outputs from autonomous vehicles being naturally temporal in nature, there is still limited exploration of exploiting point cloud sequences for 3D semantic segmentation.  In this paper we propose a novel Sparse Temporal Local Attention (STELA) module which aggregates intermediate features from a local neighbourhood in previous point cloud frames to provide a rich temporal context to the decoder.  Using the sparse local neighbourhood enables our approach to gather features more flexibly than those which directly match point features, and more efficiently than those which perform expensive global attention over the whole point cloud frame.  We achieve a competitive mIoU of 64.3\% on the SemanticKitti dataset, and demonstrate significant improvement over the single-frame baseline in our ablation studies. 
\end{abstract}

%% file: Chapters/01_Introduction.tex
\section{Introduction}
Semantic scene understanding is an essential task for the operation of autonomous vehicles.  A fine-grained understanding of the environment allows an autonomous vehicle to distinguish between drivable and non-drivable surfaces, identify moving objects in a scene, and leverage additional context for improved place recognition \cite{vidanapathirana2021locus} or global data association for Simultaneous Localization and Mapping (SLAM) \cite{park2021elasticity}. Conversely, a poor understanding of the vehicle's surroundings can negatively impact its ability to make informed decisions and can lead to fatal and other unwanted accidents.

3D point clouds collected by either LiDAR or depth cameras are a common perception modality for many autonomous vehicles, providing better geometric understanding of the surrounding environment compared to a 2D camera.  With the release of several large-scale annotated datasets in recent years \cite{behley2019semantickitti,caesar2020nuscenes}, the task of semantic segmentation of point clouds has gained in popularity.  However, despite a number of successful recent approaches exploiting sequential data from 2D video streams for improved segmentation performance \cite{hu2020temporally,li2018low,paul2020efficient,zhu2019improving,jain2019accel}, there has been limited exploration into leveraging temporal priors for point cloud segmentation.  Existing approaches either calculate strict correspondences between point features across frames \cite{caoasap} or perform global attention \cite{shi2020spsequencenet} between whole point clouds.  In the case of the former, a breakdown of nearest-point matching due to displacement between adjacent point clouds can result in the mis-matching of features across frames.  In the case of the latter, the computational cost of performing global attention scales rapidly with the size of the point cloud and also risks introducing false correlations between superficially similar but distant regions of the scene \cite{paul2021local}.  

We believe that effective leverage of point cloud sequences should gather features from a local region of the point cloud, allowing for more flexible aggregation of past information than strict correspondences while avoiding the pitfalls of global attention computations.  To accomplish this we introduce a Sparse Temporal Local Attention (STELA) module as seen in Figure \ref{fig:STELA}.  This module decomposes the sparse 3D voxel features in the network backbone into pseudo-point clouds as shown in Figure \ref{fig:decomposition}, composed of the non-empty voxel features and their corresponding grid co-ordinates.  The grid co-ordinates are used to find the spatially local neighbourhood between the past and present point cloud frames, over which we use cross-attention to aggregate salient features to provide useful temporal context to the decoder.  By constraining our cross-attention to local neighbourhoods, we focus the search for correlated past features to the most relevant region of the point cloud while also significantly reducing the computational cost involved.  In summary our contributions are: %

\begin{itemize}
    \item We introduce the Sparse Temporal Local Attention (STELA) module which aggregates local features from previous point cloud frames to improve segmentation performance. 
    \item We investigate the impact of the local neighbourhood size and sequence length on downstream performance when aggregating features from previous point cloud frames.
    \item We achieve competitive results on the SemanticKitti dataset, and demonstrate a significant boost in performance over the single-frame baseline. 
\end{itemize}

\begin{figure*}[ht]
    \centering
    \includegraphics[width=\linewidth]{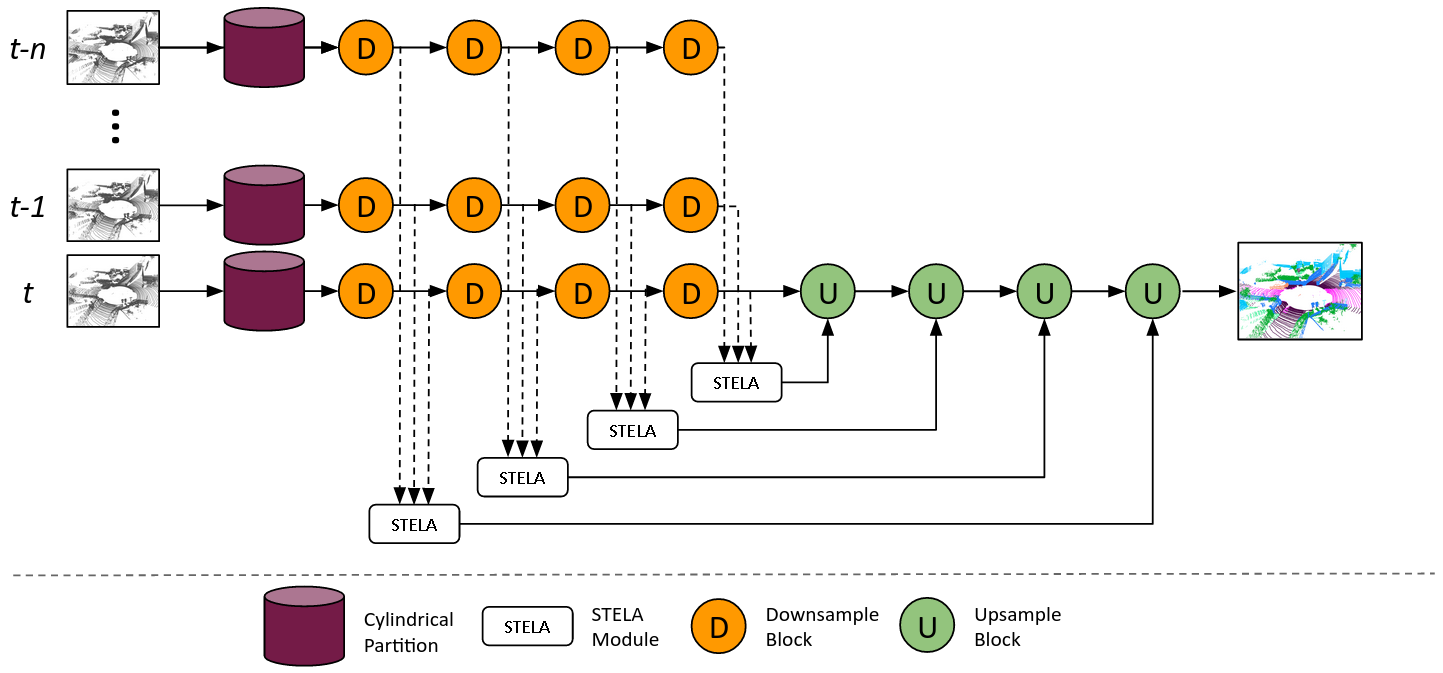}
    \caption{Our overall network architecture.  The input point clouds are projected into the cylindrical partitions as the input to a 3D U-Net style architecture.  At each skip layer, the intermediate features are fed into the STELA module which provides the input to the decoder.}
    \label{fig:architecture}
\end{figure*}

%% file: Chapters/02_Related_Works.tex
\section{Related Works}

\subsection{Semantic Segmentation in Static Point Clouds}
Previous work in semantic segmentation for static point clouds can broadly be classified into three categories: Voxelisation, 2D projection and direct point-based methods.  Voxelisation based approaches \cite{riegler2017octnet,tchapmi2017segcloud,wang2019voxsegnet,huang2016point,zhu2020cylindrical,graham20183d} partition the point cloud into a 3D tensor, over which 3D convolutions can be used to extract and classify point features.  However, the large size and sparsity of these voxelisations can present issues with memory consumption as the number of voxels increases.  To resolve this, cylindrical partitioning \cite{zhu2020cylindrical} and sparse convolutions \cite{graham20183d} have been used to reduce the sparsity of the partition and restrict operations to the active regions of the tensor.

2D projection based methods project the 3D point cloud onto a 2D plane by using the spherical projection \cite{xu2020squeezesegv3,milioto2019rangenet++,wu2018squeezeseg,wu2019squeezesegv2,cortinhal2020salsanext}, bird's-eye projection \cite{zhang2020polarnet,aksoy2019salsanet} or a fused approach \cite{alnaggar2021multi,wang2018fusing}.  This family of approaches tends to reach higher segmentation speed due to the efficient use of 2D convolutions, but suffers from the loss of geometric information in the point cloud during projection from 3D to 2D.

Direct point-based methods process the point cloud directly instead of projecting to a tensor representation.  The majority of approaches in this category are based on the PointNet architecture \cite{qi2017pointnet,pointnet++} which extracts point-wise features using a multi-layer perceptron (MLP) for segmentation.  Recent approaches have focused on extracting the local relationships between points through improved grouping \cite{engelmann2018know} or crafting convolutional operations to take place over the point cloud \cite{groh2018flex,tatarchenko2018tangent,hua2018pointwise}.

\subsection{Semantic Segmentation in Point Cloud Sequences}
Despite significant recent gains made in video semantic segmentation by incorporating sequential data \cite{hu2020temporally,li2018low,paul2020efficient,zhu2019improving,jain2019accel}, few approaches for point cloud segmentation have taken advantage of this temporal signal.  MeteorNet \cite{liu2019meteornet} and ASAP-Net \cite{caoasap} are point-based methods which use 3D flow and nearest-centre matching respectively to identify point features from past point cloud frames to be fused into the present timestep.  SpSequenceNet \cite{shi2020spsequencenet} is a voxelisation method which performs global attention to fuse highly correlated features between point cloud frames.  However, MeteorNet and ASAP-Net's reliance on flow and centre matching to calculate strict correspondence leads to issues when these priors fail, while SpSequenceNet's use of global attention is computationally expensive and introduces the possibility of finding false correlations between similar instances or classes present in different areas of the scene \cite{paul2021local}.

\begin{figure*}[ht]
    \centering
    \includegraphics[width=\linewidth]{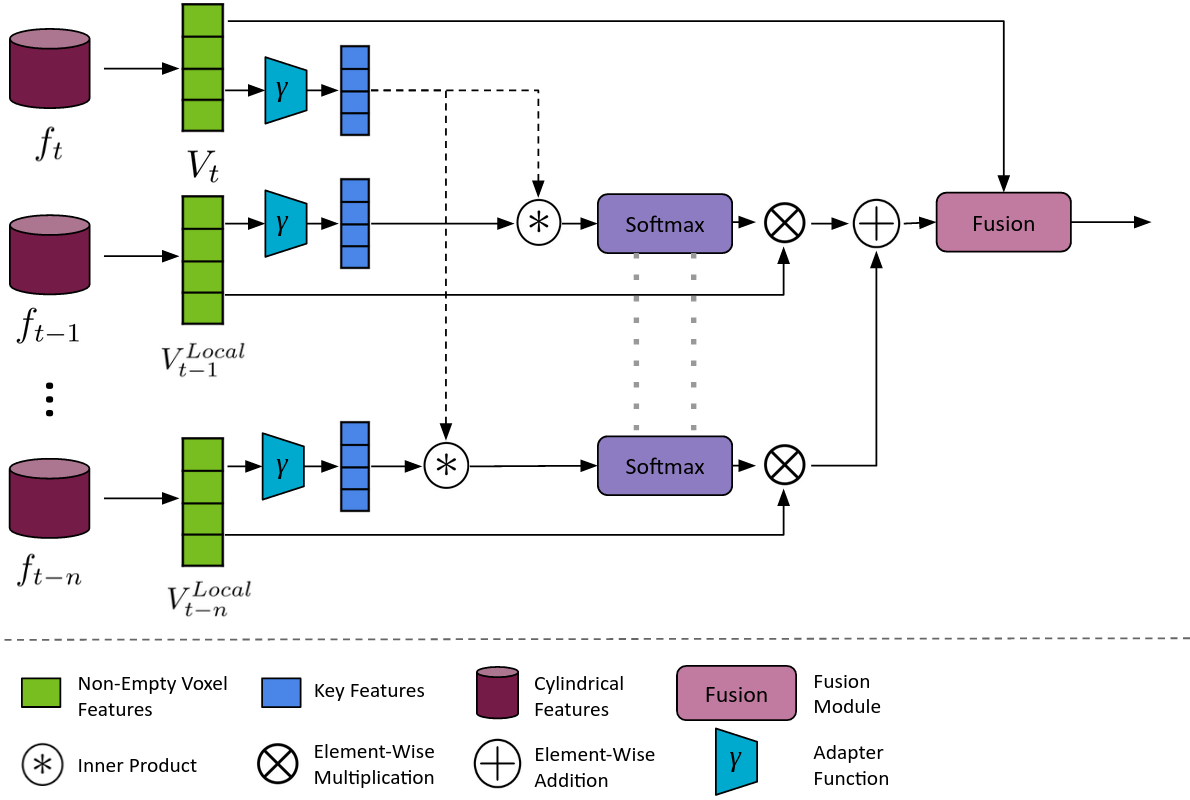}
    \caption{The STELA module.  We convert the cylindrically partitioned features into a pseudo point-cloud composed of the features from its non-empty voxels, and find the correlations with features in a local neighbourhood in previous point cloud frames.  Features with high correlation are gathered from the memory and fused with features from the present point cloud frame.}
    \label{fig:STELA}
\end{figure*}
\subsection{Temporal Cross-Attention}
Cross-attention is a variant of the attention mechanisms which are used for tasks such as sensor fusion \cite{mohla2020fusatnet,pande2021adaptive,zhang2020hybrid} and image captioning \cite{wei2020multi,lee2018stacked} where it is important to find correlations between modalities which provide different views of the same target.  We define temporal cross-attention as an emerging family of methods \cite{shi2020spsequencenet,caoasap,paul2021local,bertasius2021space} in segmentation and classification tasks which leverage the attention mechanism by treating subsequent point cloud frames of the input as different `views' of a scene.  One of the advantages of employing temporal cross-attention is the ability to focus the search for correlations to a local neighbourhood, as the temporally smooth nature of subsequent point cloud frames suggests that the most salient regions of the input for finding strong correlations will occupy a similar spatial location between point cloud frames.  This prior allows for the construction of a local neighbourhood for the attention mechanism which avoids the pitfalls of mis-matching features when strict correlation between features across point cloud frames is used \cite{caoasap}, significantly reducing the computational cost and chance of false correlations compared to global attention \cite{shi2020spsequencenet}.

%% file: Chapters/03_Methodology.tex
\section{Methodology}

In this section we describe the learning framework, details of the architecture and loss function, and other modules used to implement our proposed sparse attention mechanism for point cloud sequences.  
\subsection{Framework Overview}
Strong correlations exist in the semantic properties of a scene between frames, so it is natural to exploit this temporal consistency instead of processing each point cloud frame independently.  However, point clouds for outdoor scenes contain several properties which impose challenges to drawing on these temporal correlations: they are sparse, have variable input size, and points relating to the same object in the world may be occluded in subsequent frames due to changes in the sensor's viewpoint.  In order to address these issues we propose a novel Sparse Temporal Local Attention module (STELA), as depicted in Figure \ref{fig:STELA}.  Section \ref{Method:Partitioning} outlines the projection of the point cloud sequence into sparse cylindrical partitions.  Section \ref{Method:Local} outlines the distinction between performing local and global attention for feature correlation, and Section \ref{Method:Neighbourhood} describes how we find the sparse local neighbourhood between point cloud frames.  Finally Sections \ref{Method:Aggregation} and \ref{Method:Fusion} detail how we use attention to find correlations over these local neighbourhoods and produce an aggregated embedding of features from previous point cloud frames, and fuse these features with the present to provide an input for the decoder.

\subsection{Partitioning the Point Cloud}
\label{Method:Partitioning}
\subsubsection{Cylindrical Representation}
Outdoor point clouds vary significantly in density based on distance to the LiDAR sensor.  In order to achieve a lower proportion of empty voxels in the partitioned point cloud, we employ a cylindrical projection which transforms our input points from co-ordinate system $\left(x,y,z\right)$ to $\left(p, \theta, z\right)$ where $p, \theta$ represent the range (distance to the origin along the x-y plane) and azimuth (angle from x-axis to y-axis) of a point respectively.  This style of partition is suited to outdoor scenes, since the increasing grid size compensates for the growth in sparsity as the range from the sensor increases.  This leads to a more balanced distribution of points in the input partition and a more even proportion of non-empty voxels \cite{zhu2020cylindrical}.  We then use a combination of the original $x,y,z$ co-ordinates, the newly found $p,\theta$ co-ordinates and the LiDAR intensity as the input for a four-layer MLP to generate the features from each point, and assign them to the partition based on their cylindrical co-ordinates.  After these steps, we unroll the cylinder to get the 3D feature embedding $F \in \mathbb{R}^{D \times H \times W \times L}$, where  $D$ denotes the feature dimension and $H,W,L$ the range, azimuth and height respectively.

\subsubsection{Sparse Feature Maps as Pseudo-Point Clouds}

\begin{figure}[ht]
    \centering
    \includegraphics[width=0.6\linewidth]{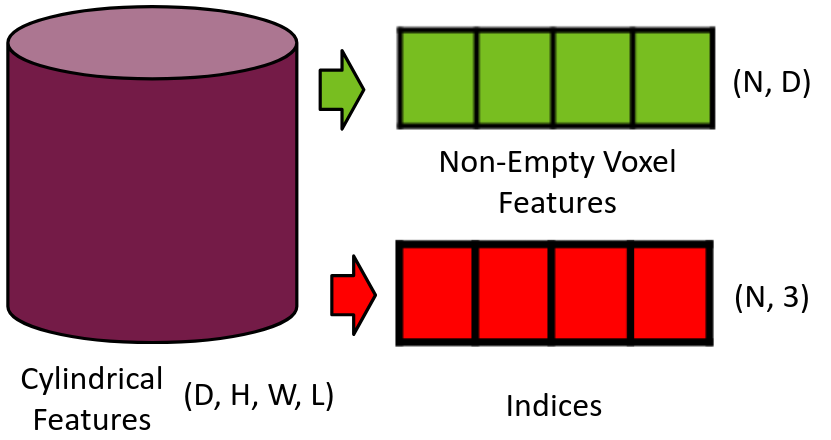}
    \caption{Decomposition of the cylindrical partition into a pseudo-point cloud.  Given $N$ non-empty voxels in the partition we can craft a $N \times D$ feature tensor where $D$ is the size of the feature dimension, and an $N \times 3$ index tensor containing the co-ordinates of these features in the partition.}
    \label{fig:decomposition}
\end{figure}

Inspired by sparse convolutional methods, given an input sparse 3D feature map $F \in \mathbb{R}^{D\times H\times W\times L}$ we decompose the useful information in the embedding into two tensors $V \in \mathbb{R}^{N\times D}$ and $I \in \mathbb{R}^{N\times 3}$.  Here $V$ and $I$ represent the features and grid indexes of the $N$ non-empty voxels in the embedding as illustrated in Figure \ref{fig:decomposition}.  We use these features and grid indexes to calculate our keys and local neighbourhood for the STELA module as described below, and convert them back to sparse 3D partitions between modules for 3D convolution.

\subsection{Local vs. Global Attention}
\label{Method:Local}
When aggregating past features, a global matching can be problematic for several reasons.  Large 3D feature maps can lead to unreasonable memory consumption and degradation of inference time, scaling rapidly as the size of the partition increases.  In addition, matching features at distant locations has the potential to introduce false correlations between similar instances or classes in a scene \cite{paul2021local}.  For these reasons we spatially align the point clouds from previous frames with the present using the local sensor odometry and constrain our attention mechanism for each voxel to a spatially local neighbourhood. %

\subsection{Sparse Local Neighbourhood}
\label{Method:Neighbourhood}
Unlike methods applying local attention in the video domain \cite{paul2021local}, given the sparse nature of the 3D partitions used to represent outdoor point cloud scenes, simply using the spatially adjacent voxels for our local neighbourhood would result in redundant calculations over a large number of empty voxels during training and inference.  In this section, we introduce a novel method of finding the local neighbourhood given a sparse input in order to avoid these redundant correlations.

Let $f_t = \left\{V_t, I_t\right\}$ and $f_{t-n}=\left\{V_{t-n}, I_{t-n}\right\}$ represent the decomposed features for the point cloud embeddings at times $t$ and $t-n$.  To find our local neighbourhood $L_{t-n}\left(i\right)$ for a given non-empty voxel $i \in \left\{1, \dots, N\right\}$ in $f_t$, we find the $k$ spatially closest non-empty voxels in $f_{t-n}$ using K-Nearest Neighbours,

\begin{equation}
    L_{t-n}\left(i\right) = \textup{KNN}\left(I_t\left(i\right), I_{t-n}, k\right),
\end{equation}

with $L_{t-n}\left(i\right)$ representing the set of non-empty voxels in the local neighbourhood for past point cloud frame $t-n$, corresponding to a non-empty voxel in the current frame $t$.  Similarly to $f_{t-n}$, each voxel in $L_{t-n}$ has an associated feature and grid index such that $L_{t-n} = \left\{V_{t-n}^L, I_{t-n}^L\right\}$, where $V_{t-n}^L \in \mathbb{R}^{k \times D}$ and $I_{t-n}^L \in \mathbb{R}^{k \times 3}$.

\subsection{Keys Matching and Memory Reading}
\label{Method:Aggregation}
In order to find the correlations between past and present point cloud frames, we first pass our voxel features through an adapter function $\gamma\left(\cdot\right)$ in order to create a lower dimensional key embedding with feature dimensionality $D_K$.  For this adapter function we have chosen a two-layer MLP.  Correlations can then be found by calculating the scaled dot product,

\begin{equation}
    C_{t-n}\left(i\right) = \frac{\gamma\left(V_t\left(i\right)\right)\cdot \gamma\left(V_{t-n}^{Local}\left(i\right)\right)^\intercal}{\sqrt{D_K}}, 
\end{equation}

where $C_{t-n}\left(i\right) \in \mathbb{R}^{N \times k}$ represents the correlation scores between the non-empty voxels in the point cloud frame at time $t$ and its local neighbourhood in the frame at time $t-n$.  We can then convert this to a mask $P_{t-n}$ for the past features by taking the SoftMax over the correlations for all point cloud frames,

\begin{equation}
    P_{t-n}\left(i, j\right) = \frac{\exp\left(C_{t-n}\left(i,j\right)\right)}{\sum_{j,n}\exp\left(C_{t-n}\left(i,j\right)\right)} ,
\end{equation}

where $j \in \left\{1, \dots, k\right\}$ represents the range over the local neighbourhood.  We can now aggregate past features to create our memory tensor $\widetilde{V}_M$ as follows,

\begin{equation}
    \widetilde{V}_M\left(i\right) = \sum_j\sum_n P_{t-n}\left(i,j\right)V_{t-n}^{Local}\left(i,j\right).
\end{equation}

This memory tensor represents a weighted sum of all past features over the $n$ past point cloud frames and local neighbourhood $k$.  By gathering these features we are able to provide key temporal context to the decoder, improving our downstream performance.

\subsection{Memory Fusion}
\label{Method:Fusion}
When fusing the gathered past and present features, ideally the network should be able to identify when to trust the past features and when circumstances such as objects entering the scene, occlusions, or irregular motions may impact the trustworthiness of the gathered features.  To do this we adapt the fusion module introduced by \cite{paul2021local} to the point cloud domain, concatenating the inputs to create a sigmoid-gating function to weigh the relative inputs of the past and present features.  Formally, this function can be expressed as

\begin{equation}
    \begin{split}
    \widetilde{V}_t & = \sigma\left(G_t\left(V_t, \widetilde{V}_M \right)\right)V_t \\
    & + \sigma\left(G_M\left(V_t, \widetilde{V}_M \right)\right)\widetilde{V}_M,
    \end{split}
\end{equation}

where $\sigma$ represents the sigmoid activation function and $G_t, G_M$ are the learned linear layers for masking the inputs of the present and memory features respectively.

%% file: Chapters/04_Experiments.tex
\section{Experimental Results}
\begin{table*}[t]
\caption{Results of our proposed method and state-of-the-art LiDAR Segmentation methods on SemanticKITTI test set.}
\label{tab:SemKitti}
\centering
\begin{adjustbox}{width=\textwidth}
\begin{tabular}{|c|c|c|c|c|c|c|c|c|c|c|c|c|c|c|c|c|c|c|c|c|}
\hline
\textbf{Methods} & \textbf{mIoU} & \rotatebox{90}{car} &  \rotatebox{90}{bicycle} & \rotatebox{90}{motorcycle} & \rotatebox{90}{truck} & \rotatebox{90}{other-vehicle} & \rotatebox{90}{person} & \rotatebox{90}{bicyclist} & \rotatebox{90}{motorcyclist} & \rotatebox{90}{road} & \rotatebox{90}{parking} & \rotatebox{90}{sidewalk} & \rotatebox{90}{other-ground} &
\rotatebox{90}{building} & \rotatebox{90}{fence} & \rotatebox{90}{vegetation} & \rotatebox{90}{trunk} & \rotatebox{90}{terrain} & \rotatebox{90}{pole} & \rotatebox{90}{traffic} \\
\hline
\hline
Darknet53~\cite{behley2019semantickitti} & 49.9 & 86.4 & 24.5 & 32.7 & 25.5 & 22.6 & 36.2 & 33.6 & 4.7 & 91.8 & 64.8 & 74.6 & {27.9} & 84.1 & 55.0 & 78.3 & 50.1 & 64.0 & 38.9 & 52.2 \\
\hline
RangeNet++~\cite{milioto2019rangenet++} & 52.2 & 91.4 & 25.7 & 34.4 & 25.7 & 23.0 & 38.3 &  38.8 & 4.8 & {91.8} & {65.0} & 75.2 & 27.8 & 87.4 & 58.6 & 80.5 & 55.1 & 64.6 & 47.9 & 55.9 \\
\hline 
RandLA-Net~\cite{hu2020randla} & 53.9 & 90.7 & 73.7 & 60.3 & 20.4 & 86.9 & 94.2 & 40.1 & 26.0 & 25.8 & 38.9 & 81.4 & 61.3 & 66.8 & 49.2 & 48.2 & 7.2 & 56.3 & 49.2 & 47.7 \\
\hline
PolarNet~\cite{zhang2020polarnet} & 54.3 & 93.8 & 40.3 & 30.1 & 22.9 & 28.5 & 43.2 & 40.2 & 5.6 & 90.8 & 61.7 & 74.4 & 21.7 & {90.0} & 61.3 & 84.0 & 65.5 & 67.8 & 51.8 & 57.5  \\
\hline
SqueezeSegv3~\cite{xu2020squeezesegv3} & 55.9 & 92.5 & 38.7 & 36.5 & 29.6 & 33.0 & 45.6 & 46.2 & {20.1} & 91.7 & 63.4 & 74.8 & 26.4 & 89.0 & 59.4 & 82.0 & 58.7 & 65.4 & 49.6 & 58.9  \\
\hline
KPConv~\cite{thomas2019kpconv} &58.8& 96.0&32.0 & 42.5 & 33.4&44.3&61.5 & 61.6 & 11.8 & 88.8 & 61.3&  72.7&31.6& \bf{95.0} & 64.2 & 84.8 & 69.2 & 69.1 & 56.4 & 47.4 \\
\hline
Salsanext~\cite{cortinhal2020salsanext} & 59.5 & 91.9 & 48.3 & 38.6 & 38.9 & 31.9 & 60.2 & 59.0 & 19.4 & 91.7 & 63.7 & 75.8 & 29.1 & 90.2 & 64.2 & 81.8 & 63.6 & 66.5 & 54.3 & 62.1 \\
\hline
FusionNet~\cite{zhang2020deep} & 61.3 & 95.3 & 47.5 & 37.7 & 41.8 & 34.5 & 59.5 & 56.8 & 11.9 & \textbf{91.8} & \textbf{68.8} & \textbf{77.1} & 30.8 & 92.5 & \bf{69.4} & 84.5 & 69.8 & 68.5&60.4 & \bf{66.5} \\ 
\hline
SPVNAS~\cite{tang2020searching} & 66.4 & - & - & - & - & - & - & - & - & - & - & - & - & - & - & - & - & - & - & - \\
\hline
Asymm-Cylinder \cite{zhu2020cylindrical} & \bf{67.8} & \bf{97.1} & \bf{67.6} & \bf{64.0} & \textbf{59.0} & \bf{58.6} & \bf{73.9} & \bf{67.9} & \bf{36.0} & {91.4} & {65.1} & {75.5} & \bf{32.3} & {91.0} & 66.5 & \bf{85.4} & 71.8 & \bf{68.5} & 62.6 & {65.6}  \\
 \hline
 \hline 
Ours (STELA) & 64.3 &  96.9 & 62.2 & 60.6 & 49.4 & 53.5 & 71.9 & 68.8 & 9.3 & 90.9 & 65.3 & 74.6 & 13.3 & 90.3 & 62.7 & 85.1 & \bf{71.9} & 67.7 & \bf{63.1} & 64.4 \\
\hline 
\end{tabular}
\end{adjustbox}
\end{table*}

In this section first we introduce the datasets used in our experiments and the details surrounding our choice of network architecture and hyper-parameters.  Next, we report results on ablation studies surrounding the impact of using frame alignment, the size of our local neighbourhood and the number of past frames used.  Finally, we compare our network trained on the full SemanticKitti dataset to state-of-the-art results.

\subsection{Dataset and Metric}
\subsubsection{SemanticKitti}
SemanticKitti \cite{behley2019semantickitti} is a large-scale driving dataset for point cloud segmentation, containing labels for both semantic and panoptic segmentation.  The dataset consists of 22 sequences, splitting sequences 00 to 10 as training set (where sequence 08 is used as the validation set), and sequences 11 to 21 as test set leading to a total of 43,351 frames overall.  After merging moving and non-moving labels there are 19 foreground classes used for training and testing, with a small number of ignore labels.

\subsubsection{Tiny SemanticKitti} 
Tiny SemanticKitti is a subset of the SemanticKitti dataset we introduce in this paper, with a train set consisting of sequence 0 of the SemanticKitti dataset and a validation set consisting of every 10th frame in sequence 8.  We use Tiny SemanticKitti to perform our ablation studies, but report the final results in Table \ref{tab:SemKitti} by training on the full SemanticKitti train set and testing with the full SemanticKitti test set.

\subsubsection{Evaluation Metric}
To evaluate our proposed method we use the mean intersection over union metric (mIoU).  This metric can be formulated as
\begin{equation}
    \textup{mIoU} = \sum_c \frac{\textup{TP}_c}{\textup{TP}_c+\textup{FP}_c+\textup{FN}_c}.
\end{equation}
Where $\textup{TP}_c,\textup{FP}_c,\textup{FN}_c$ represent the true positives, false positives and false negative predictions for class $c$.

\subsection{Implementation Details}
To show the importance of temporal information, we used a basic encoder-decoder architecture along with our STELA module as seen in Figure \ref{fig:architecture}.  The output from each downsample block is updated using a STELA module before being passed to its corresponding upsample block through a skip connection.  This architecture is adapted from \cite{zhu2020cylindrical}, using cylindrical voxel co-ordinates and asymmetrical convolution in the downsample and upsample blocks.  However, we do not adopt the additional point-based feature branch used in their approach.  We use an input voxel size of $240 \times 180 \times 16$ for the ablation studies and $480 \times 360 \times 32$ for our results compared to the state-of-the-art.  We train with a batch size of 4 on 4 Tesla-P100 GPUs.

Our training takes place in three stages.  Firstly, we train a model without the STELA module to perform single-frame segmentation.  Next, we insert the STELA module into each skip layer of the network and freeze the encoder and decoder in order to initialize the weights of the module.  Finally, we unfreeze the network and train end-to-end for 40 epochs or until the model reaches convergence.  We use a fixed learning rate of $1e^{-3}$ for the first two stages of training, and a learning rate of $1e^{-5}$ decaying on plateau for the final step.

For our loss function, we follow existing methods \cite{cortinhal2020salsanext,zhu2020cylindrical} and use a combination of the weighted cross-entropy and lovasz-softmax \cite{berman2018lovasz} losses to maximise the accuracy and IoU score of the network output respectively.

\subsection{Ablation Studies}
\subsubsection{Aligning Inputs}
\begin{table}[ht]
    \begin{center}
    \begin{tabular}{|c|c|c|c|}
        \hline 
        Method & Aligned & mIoU & $\Delta \%$ \\
        \hline \hline 
         Baseline & - & 52.57 & - \\
         STELA & \xmark & 52.87 & 0.3\% \\
         STELA & \cmark & \textbf{53.97} & \textbf{1.4}\% \\
         \hline 
    \end{tabular}
    \end{center}
    \caption{Impact of aligning point clouds with current frame on segmentation performance on Tiny SemanticKitti.  We use a local neighbourhood size and sequencehttps://www.overleaf.com/project/601ccddf620035b0019fbed0 length of $k=16$ and $n=2$ for this ablation.}
    \label{tab:alignment}
\end{table}

Table \ref{tab:alignment} shows the result of aligning the point clouds into the sensor reference frame at time $t$.  Baseline refers to the performance of the network architecture without the STELA module, taking a single frame as an input.  Aligning the point clouds results in a $1.1\%$ boost in performance from the STELA module, though a small improvement is still observed when alignment does not take place.

\subsubsection{Local Neighbourhood Size}
\begin{table}[ht]
    \begin{center}
    \begin{tabular}{|c|c|c|}
        \hline 
        Local Neighbourhood Size & mIoU & $\Delta \%$ \\
        \hline \hline 
        Baseline & 52.57 & -\\
         4 & 53.68 & 1.11\% \\
         8 & 53.8 & 1.23\% \\
         16  & \textbf{53.97} & \textbf{1.4}\% \\
         32 & 53.6 & 1.03\% \\
         64 & 53.64 & 1.04\% \\
         \hline 
    \end{tabular}
    \end{center}
    \caption{Impact of local neighbourhood size on segmentation performance on Tiny SemanticKitti.  We use a sequence length of $n=2$ for this ablation.}
    \label{tab:neighbourhood}
\end{table}
Table \ref{tab:neighbourhood} shows the effect of the size of the local neighbourhood, measured by the number of $k$ nearest non-empty neighbours.  We can see that increasing the size of the local neighbourhood causes incremental improvements in the downstream performance up to $k=16$, at which point performance diminishes as the size of the neighbourhood is increased.  This result suggests a confirmation of our hypothesis that by focusing on the local neighbourhood when calculating attention, we capture the salient region of the point cloud for correlations while saving computational resources by disregarding the rest of the frame.

\subsubsection{Sequence Length}
\begin{table}[ht]
    \begin{center}
    \begin{tabular}{|c|c|c|}
        \hline 
        Past Frames & mIoU  & $\Delta \%$\\
        \hline \hline 
         Baseline & 52.57 & - \\
         1 & 52.89 & 0.32\% \\
         2  & \textbf{53.97} & \textbf{1.4\%} \\
         3 & 53.75 & 1.18\% \\
         4 & 53.39 & 0.82\% \\
         \hline 
    \end{tabular}
    \end{center}
    \caption{Impact of past frames used on segmentation performance on Tiny SemanticKitti.  We use a local neighbourhood size of $k=16$ for this ablation.}
    \label{tab:receptive}
\end{table}
Table \ref{tab:receptive} shows the effect of changing the number of frames used in the sequence input to the STELA module.  We achieve our strongest result using a sequence length containing two past point cloud frames (total sequence length three), after which performance begins to decline.  This is likely since as the time between frames increases, the increasing differences between present and past diminishes the usefulness of the prior features.  However, our result demonstrates that there is benefit in searching for correlations over more than a single past frame.

\subsection{Comparison to State-of-the-Art}
Table \ref{tab:SemKitti} compares our result to the state-of-the-art on the test split of the SemanticKITTI dataset.  We achieve competitive results with the state-of-art, outperforming to the best of our knowledge all but two other methods on the SemanticKitti test set.  We believe this strong overall result combined with our demonstrated improvements on the Tiny SemanticKitti validation set confirm the benefits of our approach over the single-frame baseline.

%% file: Chapters/05_Conclusion.tex
\section{Conclusion}
In this paper we propose the novel attention module STELA to aggregate sequential information from temporally adjacent point clouds.  By restricting our attention mechanism to the local spatiotemporal neighbourhood, we improve performance by focusing on the most relevant past features while also reducing the computational expense of finding correlations.  For future work we wish to investigate improving our attention mechanism by incorporating stronger geometric priors such as a positional embedding into the keys, and integrating learned odometry into the pipeline in order to reduce reliance on external priors for frame alignment.